\def\year{2020}\relax
\documentclass[letterpaper]{article} 
\usepackage{aaai20}  
\usepackage{times}  
\usepackage{helvet} 
\usepackage{courier}  
\usepackage[hyphens]{url}  
\usepackage{graphicx} 
\usepackage{graphicx}  
\usepackage{algorithm}
\usepackage{algpseudocode}

\usepackage{subcaption}

\usepackage{amsfonts}
\usepackage{amsmath}
\usepackage{booktabs}
\usepackage{amsthm}
\theoremstyle{plain}
\newtheorem*{rem}{Remark}
\newtheorem*{defn}{Definition}

\title{Justification-Based Reliability in Machine Learning\thanks{Extended version of paper accepted at AAAI 2020}}

\author{\Large \textbf{Nurali Virani\thanks{All authors contributed equally.}, Naresh Iyer, Zhaoyuan Yang}\\ 
	GE Research\\
	1 Research Circle, Niskayuna, NY 12309\\
	\{nurali.virani, iyerna, zhaoyuan.yang\}@ge.com 
}
\begin{document}
\maketitle
%
%
%
%
%
%
\begin{abstract}
With the advent of Deep Learning, the field of machine learning (ML) has surpassed human-level performance on diverse classification tasks. At the same time, there is a stark need to characterize and quantify reliability of a model's prediction on individual samples. This is especially true in application of such models in safety-critical domains of industrial control and healthcare. To address this need, we link the question of reliability of a model's individual prediction to the epistemic uncertainty of the model's prediction. More specifically, we extend the theory of Justified True Belief (JTB) in epistemology, created to study the validity and limits of human-acquired knowledge, towards characterizing the validity and limits of knowledge in supervised classifiers. We present an analysis of neural network classifiers linking the reliability of its prediction on an input to characteristics of the support gathered from the input and latent spaces of the network. We hypothesize that the JTB analysis exposes the epistemic uncertainty (or ignorance) of a model with respect to its inference, thereby allowing for the inference to be only as strong as the justification permits. We explore various forms of support (for e.g., $k$-nearest neighbors~($k$-NN) and $\ell_p$-norm based) generated for an input, using the training data to construct a justification for the prediction with that input. Through experiments conducted on simulated and real datasets, we demonstrate that our approach can provide reliability for individual predictions and characterize regions where such reliability cannot be ascertained.
\end{abstract}

\section{Introduction}
Predictive and prescriptive models based on ML are at the heart of the modern digital revolution. They are applied across several domains, such as energy, finance, defense, network security, and healthcare. These models often directly influence control actions or indirectly influence decisions through recommendations. In many scenarios, especially safety-critical ones, although the average performance or accuracy of such models is important, the reliability of the model on every individual prediction is equally critical. By reliability, we mean the degree to which the result of the model can be relied on to be accurate. A model with very good average performance is still capable of causing irreparable damage due to the mistakes it makes on a single or few predictions. In this paper, we present an analysis on the reliability of individual predictions of a ML model. 

We draw inspiration from epistemology, or the theory of knowledge, which deals with validity of methods used to acquire knowledge. Specifically, we examine the applicability of Plato's classical Justified True Belief (JTB) theory in epistemology~\cite{ichikawa2001analysis}. The JTB theory of knowledge is a classical theory that attempts to provide necessary and sufficient conditions under which a person can be said to \textit{know} something. We extend this theory to ML models by firstly positing that an individual prediction from a model can be seen as a \textit{belief} equivalent to ``I \textit{believe} input x is of class y". Next, we propose that in order for such belief to be considered as \textit{knowledge}, an additional step of constructing justification for the belief is necessary. We explore formulations of JTB analysis by which ML models can construct such justifications for their beliefs. 

We contend that existing approaches tend to conflate epistemic uncertainty, which pertains to uncertainty due to information that is absent, with aleatoric uncertainty, which pertains to uncertainty arising from variability in the available information. JTB analysis tackles this fundamental confusion by not relying on confidence intervals for point-class assignments as assessments of prediction reliability, and by abstaining from making point-class predictions on inputs, if concrete evidence for such assessment is absent. We recognize that there are existing approaches that make use of thresholding on softmax layer values or distance from the decision hyperplane to abstain from making a class assignment. However, none of these existing approaches link prediction reliability to the location of the sample in the overall input space. It is becoming clearer based on recent work~\cite{meng2017magnet} that input regions, either close to the decision boundary or in regions of extrapolation, are particularly susceptible with regard to making unreliable predictions. We present a mechanism for characterizing regions in input space as region of extrapolation (i.e., ``I don't know"  or IDK), region of confusion (i.e., ``I may know"  or IMK), and region of trust (i.e., ``I know" or IK). This novel mechanism to construct justified belief with IK, IMK, and IDK directly affects the reliability of the decisions made downstream in the analytical workflows. IMK informs that support is impure and additional information is needed to improve reliability and empty neighborhood in IDK informs that input is anomalous/novel.

Additionally, it has been shown~\cite{de2016deep} that the softmax classifier can make high confidence predictions that are inaccurate even for points that are at a large distance from the decision hyperplane. JTB analysis overcomes these issues by explicitly factoring in the location of a sample in the input space, while making its class prediction, through constructing support for the prediction based on other training data points in the vicinity of the sample. Suggestions related to abstaining from making point-class predictions have been made recently~\cite{shafahi2018adversarial,rouani2019safe,miller2019not} (especially for adversarial attack detection), but we believe our approach is the first to separately characterize overlap (IMK) and extrapolation (IDK) for individual predictions from neural networks. Other related work like DeepKNN~\cite{papernot2018deep} and~\cite{jiang2018trust} collect evidence for prediction reliability by identifying nearest neighbors to the sample from the training points across all layers of a neural network; the class-membership of the neighbors are then used to assign a nonconformity or trust score to augment to the class prediction. Although our approach leverages this work, by considering the $k$-NN support operator for the JTB analysis, in addition to other forms of support like $\ell_p$-norm based and novel hybrid forms of support, our work is different as we can identify region of extrapolation or IDK regions and we provide justification for individual reliability, whereas the statistical nature of trust scores needs thresholding to determine predictions that one can likely trust.



The main objectives and contributions of this work are given as follows:
\begin{itemize}
	\item To introduce, formalize, and illustrate the notion of justified belief from epistemology for ML models to gain reliability in individual predictions.
	\item To contrast performance of Epistemic classifiers with those that rely on distance from hyperplane on real and simulated datasets.
\end{itemize} 

We next introduce Epistemic classifiers, instantiated by the application of JTB analysis to standard classifier models. We use the rest of the paper to define concepts necessary for formalizing the JTB analysis framework, describing algorithmic steps for performing JTB analysis and inference related to it, describing experiments with applying Epistemic classifiers and analysis of outcomes from it, and finishing with a discussion of potential future directions of our work and conclusions.
 
\section{Concept of Epistemic Classifiers}
\begin{figure}[!t]
	\centering
	\includegraphics[width=0.8\linewidth]{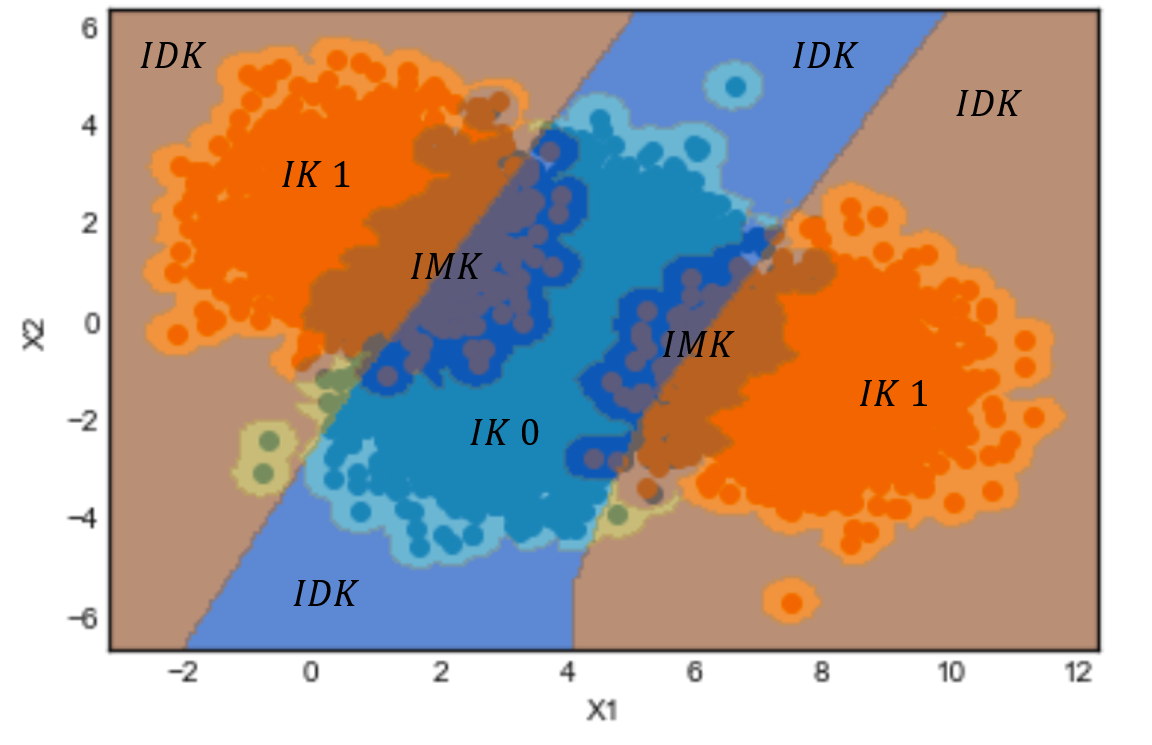}
	\caption{Illustration of region of trust (IK0, IK1), region of confusion (IMK), and region of extrapolation (IDK) for 2-D input binary classification with Epistemic classifier using neural network as base classifier}
	\label{fig:viz}
\end{figure}

Presently, a classification problem is not evaluated in terms of the true class observability permitted by the training data used to construct classifiers. In reality, the ground truth of a particular instance might not be fully observable from the available training data. As a result, reliable single-point classification for the dataset might not be possible. We show how Epistemic classifiers can successfully handle this issue, through defensible abstention, while traditional classifiers are forced to make point-class predictions without regard to whether or not the information space permits many of those point-class predictions. This impacts their individual prediction reliability. Epistemic classifiers attempt to characterize regions of observability and abstain from making class-assignments in regions of the input space where such characterization is found to be difficult to construct. Consider Fig.~\ref{fig:viz} illustrating the input space of a 2-D binary classification problem. The problem we are interested in is, given a test input $x$, to be able to characterize the input region for $x$ during inference and provide estimated class-prediction with additional information indicating individual reliability of the prediction. The fundamental mechanism to enable this characterization of the input space is developed by application of the theory of Justified True Belief (JTB) from field of epistemology, which is discussed next.


The JTB theory of knowledge suggests that a subject S \textit{knows} that a proposition P is true if and only if P is true, S believes that P is true, and S is justified in believing that P is true. 
One key observation is that in JTB analysis, the truth state of proposition is leveraged in the determination of knowledge of subject, which might not be practical. In fact, most of scientific knowledge is based on justified belief, where belief is a claim or hypothesis and justification is the experimental evidence or the mathematical proof leveraging existing justified beliefs to support the belief. Thus, knowledge as justified belief will be used in this research. We extend this theory of knowledge to supervised learning-based classifiers. Here belief of artificial intelligence (AI) is the same as output of the classifier model. However, we have freedom in construction of justification and strength or defensibility of justification will determine reliability of AI's knowledge.         

If observation $x$ is a camera image, proposition $P$ is that ``$x$ shows a human", AI ``believes" in $P$ as the classifier gives an output of human class, then AI \textit{knows} $P$ given $x$, iff AI is justified in having that belief. In current AI systems, \textit{cross-validation accuracy of model is high} is used as a justification to rely on model outputs, but this kind of justification using aggregate statistics fails to account for unreliability at the level of individual predictions. Thus, we introduce justification that gathers evidence from training set for an individual test input $x$ in input and hidden layers of a neural network classifier, where an unambiguous truth state in the neighborhood of $x$ in those embedded spaces provides support to the belief allowing model to declare "I know $P$". This support and justification process to characterize knowledge, which uses neighborhood to allow generalization beyond labeled instances, is discussed next.

\subsection{Neighborhood and Support}
\begin{figure}[t!]
\includegraphics[width=\linewidth]{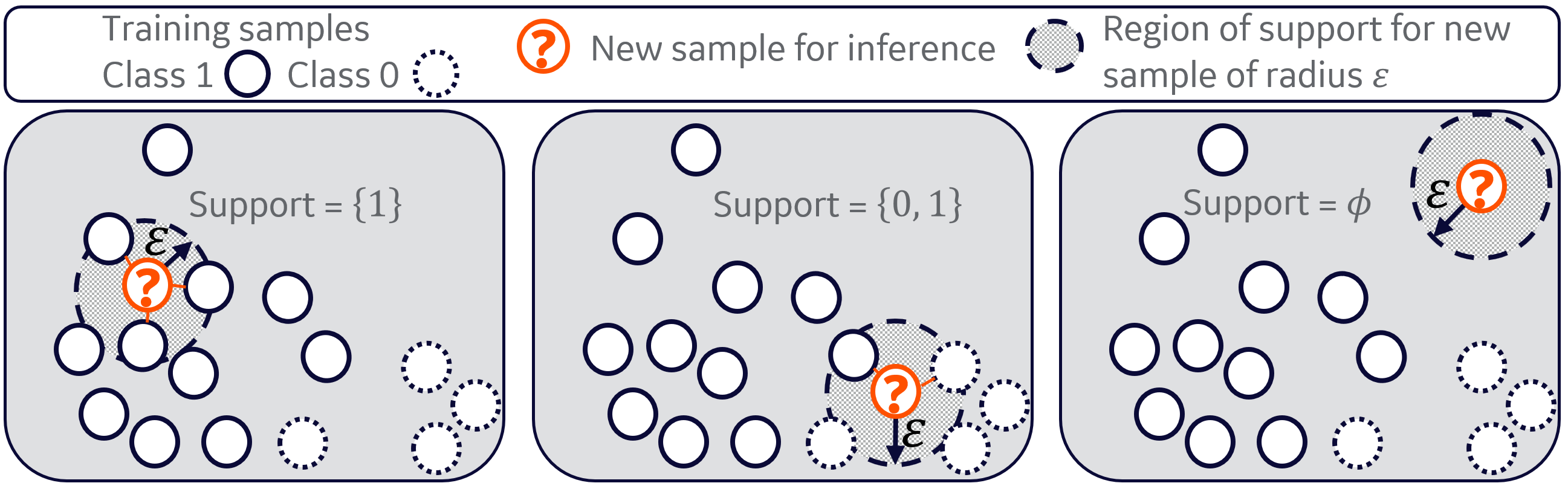}
\caption{Construction of support in some embedded space using $\varepsilon$-balls as neighborhoods}
\label{fig:support}
\end{figure}

Consider a training set with predictors $X$ and set of corresponding labels $Y$. A neighborhood operator $\mathcal{N}(x)$ for an input $x$ is a function that returns a subset of training set, i.e., $\mathcal{N}(x) \subset X$, such that each data point $z \in \mathcal{N}(x)$ is similar to $x$ using certain distance metric and selection criterion. Specific examples of neighborhood operators include $\varepsilon$-ball neighborhood $B_{\varepsilon}(x)$ which uses $\ell_2$ distance metric and criterion that $d(x, z) \leq \varepsilon$ for $\varepsilon \in \mathbb{R}^+$ and $k$-nearest neighbor neighborhood $N_{k}(x)$ which uses distance metric $d$ to identify $k \in \mathbb{Z}^+$ nearest neighbors of $x$ from the training set. Let function $f$ map from input from training data to set of labels, then given a neighborhood operator $\mathcal{N}_i(\cdot)$, the support of input $x$ in $i^{th}$ layer of neural network is defined as follows:
\begin{align}
S_i(x) &= \{f(z): z \in X,  h_i(z) \in \mathcal{N}_i(h_i(x))\},
\label{eq:support}
\end{align}
where $h_i(z)$ is the activation values of layer $i$ in neural network with $z$ as input. Thus via the neighborhood operator, the support operator is parametrized by distance bound $\varepsilon_i$ or number of neighbors $k_i$. An illustration of $\varepsilon$-ball neighborhood to construct support is shown in Fig.~\ref{fig:support}.

A $k$-NN neighborhood adapts region to look for evidence based on data density. However, this favorable adaptability gives $k$ neighbors even when the sample is extrapolating and is semantically different from all neighbors. On other hand, $\varepsilon$-ball neighborhood has fixed region in which it looks for evidence and number of evidence varies based on data density and can also lead to empty support under extrapolating regions. However, this favorable variability is expected to lead to $\varepsilon$ that might be conservative in one part of input space and loose in another part. Since, the formulation can handle general notion of neighborhood, we also propose hybrid versions of neighborhood operators: 1) H-1 -  when $\varepsilon$-ball neighborhood is empty, then use $k$-NN neighborhood, and 2) H-2 - when $\varepsilon$-ball neighborhood is non-empty, compute union with $k$-NN neighborhood. Note that when $\varepsilon$-ball neighborhood is empty, H-1 provides more information about nearest neighbors, but it would potentially reduce robustness as those evidence might be obtained from semantically dissimilar classes. H-2 provides stronger robustness as it increases the set of evidence, but it would potentially increase impurity of evidence, i.e. data points from different classes. 
In this work, the neighborhood operators that are considered to construct support in layer $i$, given the choice of distance metric $d_i$, can be parametrized by size of ball neighborhood $\varepsilon_i$ and number of neighbors $k_i$, i.e. we can represent neighborhood operator as $\mathcal{N}_i = \mathcal{N}_{d_i, \varepsilon_i, k_i}$. This support operator is evaluated using activations over each layer of interest and then justification operator acts over these sets of support. This process of justification is explained next.     

\subsection{Justification and Knowledge}
The justification operator takes as input a collection of supports from multiple layers and produces a set of justified class assignments for an input using \eqref{eq:justification}. 

\begin{align}
J(x) = 
	\begin{cases}
	\phi & \text{if for any } S_i(x) = \phi, \\
	\bigcup_{i \in I} S_i(x) & \text{otherwise}.
	\end{cases} 
	\label{eq:justification} 
\end{align}

If any support is empty, then $J(x)$ is also empty, else we use the disjunction or union for the justification operator. Thus, the justified class assigned to input $x$ is a union of the class-assignments made by each of the supports from chosen layers. The disjunction makes for a liberal justification operator since a class that is contained even in one of the supports is carried on to the output. 

Based on the outcome of applying the justification operator $J(x)$ and prediction or belief $g(x)$ for a given input $x$, the Epistemic classifier using justified belief may assert IK, IMK, or IDK. These assignments capture the epistemic uncertainty of the classifier with respect to the input's class. The IDK assertion captures the epistemic uncertainty encountered when a sample finds no support for the belief from existing training data (e.g., when the inference involves extrapolation), i.e. $g(x) \not\subset J(x)$. The IMK assertion captures the epistemic uncertainty when a sample finds conflicting support from the training data for two or more classes which also consists of the belief, i.e., $g(x)$ is proper subset of $J(x)$. For example, in Fig.~\ref{fig:viz}, if test input $x = (2,2)$, then $J(x) = \{0, 1\}$ from support in input layer and $g(x) = \{1\}$ leading to IMK assertion. Additionally, the IK assertion is provided when the belief and justification are exactly the same $g(x) = J(x)$, e.g., when $x = (8, 0)$ in Fig.~\ref{fig:viz}. Clearly, these three degrees of epistemic uncertainty are ordered wherein IDK captures a relatively higher degree of ignorance about the sample class than the IMK and IMK has higher uncertainty than IK. Now that concept of Epistemic classifier and intuition has been explained we will define this classifier ahead. 

\begin{defn}[Epistemic classifier]
Epistemic classifier ($E$) is a function mapping from input space to output label and epistemic certainty. It is defined by a tuple $E = (g, I, \boldsymbol{\mathcal{N}})$, where~$g$ is the base neural network classifier, $I$ is the set of chosen layers, and $\boldsymbol{\mathcal{N}} = \{\mathcal{N}_1, \dots, \mathcal{N}_{|I|}\}$ is a set consisting of neighborhood operator for each layer. For a given input $x$, this classifier obtains belief as $g(x)$ with base classifier and constructs justification $J(x)$ using~\eqref{eq:justification} with neighborhood and support operators using~\eqref{eq:support} over layers in $I$. Then output $E(x) = (g(x), j \in \{\text{IK, IMK, IDK}\})$ consists of belief and assertion of IK (if $g(x) = J(x)$), IMK (if $g(x) \subset J(x)$), or IDK (if $g(x) \not\subset J(x)$) as the degree of epistemic certainty in the belief.
\end{defn}

The concept of Epistemic classifiers was covered in this section. Next we will show the training and inference algorithms as well as some theoretical analysis on choice of model parameters.
  
\section{Algorithm and Analysis} 


In order to build an Epistemic classifier, we need a trained neural network model $g$, which is the base classifier. The evidence for justification process will be based on the training set $(X, Y)$. The parameter optimization is conducted by evaluating metrics over the validation set $(X^v, Y^v)$. We demonstrate our training approach in Algorithm~\ref{alg:EClassifier}. 

\begin{algorithm}[b!]
	\caption{\textbf{--  Training Epistemic Classifiers}.}
	\label{alg:EClassifier}
	\begin{algorithmic}[1] 
		\Require training set $(X, Y)$, validation set $(X^v, Y^v)$
		\Require trained neural network $g$
		\Require set of layers $I$ for support construction
		\Require metrics for each layer $\bar{d} = \{d_1,...,d_{|I|}\}$   
		\State $\Omega = \{\}$ \Comment{Set of one search tree per layer in $I$}
		\For{each layer $i \in I$}
		\State $\Gamma_{i}^X \leftarrow$ Extract activation $h_{i}(X)$ for training set $X$ 
		\State $\Omega_i \leftarrow \text{NeighborSearchTree}(\Gamma_{i}^X, Y, d_i)$
		\State $\Omega \leftarrow \Omega \bigcup \Omega_i$
		\EndFor
		\State $\bar{\varepsilon}, \bar{k} = \text{ParameterSelection}(X^v, Y^v, \Omega, g, I, \bar{d})$
		\State \Return $g, I, \bar{d}, \bar{\varepsilon}, \bar{k}, \Omega$ 
	\end{algorithmic}
\end{algorithm}

\begin{algorithm}[b!]
	\caption{\textbf{--  Inference with Epistemic Classifiers}.}
	\label{alg:EClassifier_Inference}
	\begin{algorithmic}[1] 
		\Require Test input $x$
		\Require Epistemic classifier $G = (g, I, \boldsymbol{\mathcal{N}})$ 
		\State Compute belief as prediction $g(x)$
		\For{each layer $i \in I$}
		\State $\Gamma_i^x \leftarrow$ Extract activation $h_i(x)$ for test input $x$ 
		\State $S_i(x) \leftarrow \Psi(\Gamma_i^x, \mathcal{N}_i)$ \Comment{support of $x$ in layer $i$}
		\EndFor
		\State Get justification $J(x)$ from supports $S_i(x)$ using~\eqref{eq:justification}.
		\If {$g(x) = J(x)$}
		\State output $\leftarrow (\text{IK}, g(x))$
		\ElsIf {$g(x) \subset J(x)$} \Comment{proper subset}
		\State output $\leftarrow (\text{IMK}, g(x))$
		\Else     \Comment{implies $g(x) \not\subset J(x)$}
		\State output $\leftarrow (\text{IDK}, g(x))$	
		\EndIf
		\State \Return output
	\end{algorithmic}
\end{algorithm}

Once we obtain the base classifier $g$, we feed the training set $X$ into neural network model $g$ and extract~$\Gamma_{i}^X$, which is the activation from layer $i \in I$ from all training instances using $h_i(X)$. Recall that $h_i(\cdot)$ gives activation values of layer $i$ for a given input. NeighborSearchTree represents a function which builds a nearest neighbor search tree based on the information from training data and distance metric. ParameterSelection represents a function, which selects a set of parameters $\bar{\varepsilon} = \{\varepsilon_i: i \in I \}$ and $\bar{k} = \{k_i: i \in I \}$ for support construction. We will discuss these functions in detail next.

In NeighborSearchTree, we construct a set of ball-trees~\cite{omohundro1989five} denoted by $\Omega$ to store activation values and their corresponding labels for each layer $i$. Unlike k-d trees~\cite{bentley1975multidimensional}, ball-trees lead to $O(\log|X|)$ average case complexity for exact search even in higher dimensions, where complexity for k-d tree is linear. These ball-trees enable to efficiently conduct both nearest neighbor search and range search, i.e., find all neighbors within distance $\varepsilon_i$ from test input $z$. In our implementation, we use ball-tree method from scikit-learn~\cite{pedregosa2011scikit} to construct the neighbor search tree. 

Let us now consider a case where support is constructed only from one layer, then we study the effect of choice of~$\varepsilon$ over~$\{\text{IK, IMK, IDK}\}$ regions. Let $F_{IK}$ be fraction of samples that are asserted to be ``I know", similarly, we have $F_{IMK}$ and $F_{IDK}$ as fraction of samples for ``I may know" and ``I don't know" respectively. Then, we can make the following remarks about their behavior with $\varepsilon$:

\begin{rem}[Behavior of $\{\text{IK, IMK, IDK}\}$ with $\varepsilon$]
Note that $F_{IK} + F_{IMK} + F_{IDK} = 1$, since $\{\text{IK, IMK, IDK}\}$ are exhaustive and mutually-exclusive options for justified belief. If $\varepsilon \rightarrow 0$, then $F_{IK} = 0$, $F_{IMK} = 0$, $F_{IDK} = 1$, since very small neighborhoods will not be able to find any evidence. If $\varepsilon \rightarrow \infty$, then $F_{IK} = 0$, $F_{IMK} = 1$, $F_{IDK} = 0$, since very large neighborhoods will include training points from all classes as evidence. $F_{IMK}$ is a monotonically-increasing function of $\varepsilon$, since impure support cannot become pure or empty, but pure or empty support of a data point can become impure as size of neighborhood increases. Similarly, $F_{IDK}$ is a monotonically-decreasing function of $\varepsilon$, since empty neighborhood might get new evidence, but nonempty neighborhoods will not become empty as $\varepsilon$ increases.  
\end{rem}
\begin{figure}[!b]
	\centering
	\includegraphics[width=0.8\columnwidth]{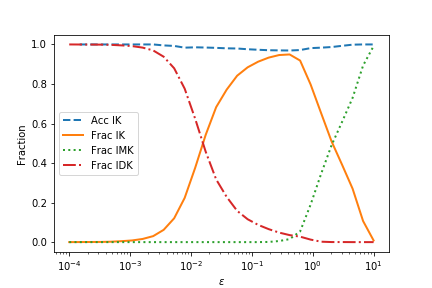}
	\caption{Illustration of behavior of fraction of IK, IMK, and IDK as well as accuracy on IK from Epistemic classifier with parameter of neighborhood operator}
	\label{fig:line}
\end{figure}

Figure~\ref{fig:line} illustrates these intuitive behaviors using a toy dataset (Gaussian distribution with some overlap) for a binary classification task. The metric $F_{IK}$ or fraction of IK samples is also called coverage. Note that accuracy for IK samples stays high over different coverage levels as desired. Thus, in this work, objective function used to choose optimal $\varepsilon$ is coverage and not accuracy. The relation of coverage with $\varepsilon$ is piece-wise constant as data points move between regions of IDK or IMK and IK. Thus, this optimization for ParameterSelection function can be accomplished with grid search or by using Bayesian optimization (BO)~\cite{snoek2012practical}, where coverage metric is evaluated over validation set.

Given $\varepsilon_i$ for one layer-case, we derive an upper bound of the $\varepsilon_{i+1}$ such that if data point $z$ is an $\varepsilon_i$-neighbor of the testing input $x$ in layer $i$, then it is guaranteed that $z$ is an $\varepsilon_{i+1}$-neighbor of the testing input $x$ in layer $i+1$. We obtain the following result:

\begin{rem}[$\varepsilon$ bound]
If~$W_i \in \mathbb{R}^{m \times n}$ is weight matrix of the layer $i$ with input dimension $m$ and output dimension $n$, $\lambda_i^*$ is the largest eigenvalue of the matrix $W_{i}W_{i}^T$, $L_i$ is the Lipschitz constant of the activation function of layer $i$, then given $\varepsilon_i$ for layer $i$, the value of $\varepsilon_{i+1}$ is bounded as follows:
$$\varepsilon_{i+1} \leq  L_i \sqrt{\lambda_i^*} \varepsilon_i.$$ 
\end{rem}
\begin{proof}
	See Supplementary material (S1).
\end{proof}

This result is used to choose conservative values of $\varepsilon$ across multiple layers by setting value of $\varepsilon_{i+1}$ at the upper bound. Note that the weight matrix of layer~$i$ leads to an affine projection of $\varepsilon_i$-ball, thus considering only scaling will lead to a loose bound in most of the directions except of the largest principal component. Thus, we also introduce a weighted distance metric based on the weight matrices of the neural network that is constructed for each layer. We define the following metric for each layer $i$, $d_i(x_a,x_b) = \sqrt{(x_a-x_b)^T D_i(x_a-x_b)}$ and $D_i$ is a symmetric positive semi-definite matrix. The following remark informs how to choose the values for $D_i$.  

\begin{rem}[Bound with weighted distance]\label{rem:distance}
Let $\widetilde{W_i} = W_1W_2...W_{i}$, $\Lambda = \text{diag}([\lambda_1, \lambda_2,...\lambda_n])$ and $V = [v_1, v_2,...v_n]$ as the eigenvalues and corresponding eigenvectors of matrix $\widetilde{W_i}^T\widetilde{W_{i}}$, and $d_0$ is $\ell_2$ distance metric in input space. If $d_0(x, z) \leq \varepsilon_0 $ then the distance function matrix $D_i$ for neighbor search in any layer $i$:
\begin{equation}
\label{eq:t}
\begin{aligned}
D_i = V\Lambda^{-1}V^T
\end{aligned}
\end{equation}
satisfies the following relation:	$$\sqrt{(h_i(x)- h_i(z))^T D_i (h_i(x)- h_i(z))} \leq \varepsilon_0$$
\end{rem}
\begin{proof}
	See supplementary material for assumptions and derivations.
\end{proof}

\begin{figure}[!b]
	\centering
	\includegraphics[width=0.8\columnwidth]{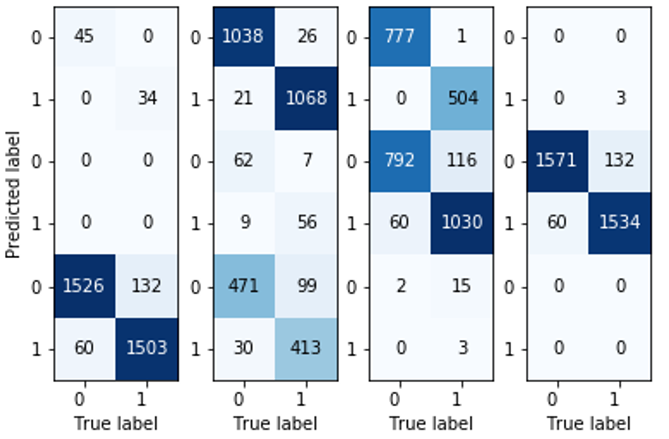}
	\caption{Effect of $\varepsilon$ ($0.001, 0.017, 0.237, 3.162$) on augmented confusion metric with noise std. of 1.0.}
	\label{fig:acm}
\end{figure}

\begin{figure*}[!t]
	\begin{subfigure}[b]{0.32\textwidth}
		\includegraphics[width=\textwidth]{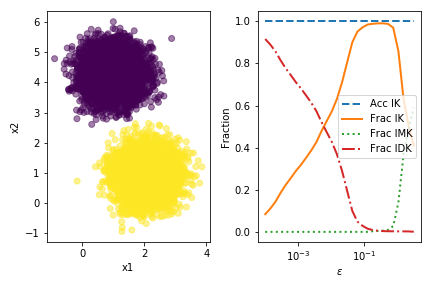}
	\end{subfigure}
	~
	\begin{subfigure}[b]{0.32\textwidth}
		\includegraphics[width=\textwidth]{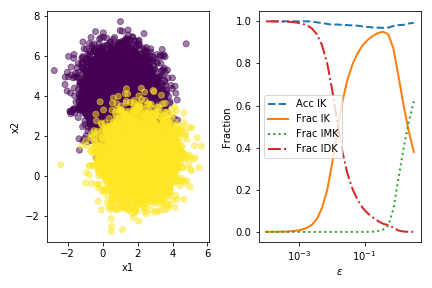}
	\end{subfigure}
	~
	\begin{subfigure}[b]{0.32\textwidth}
		\includegraphics[width=\textwidth]{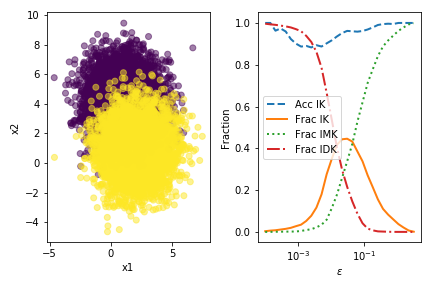}
	\end{subfigure}
	\caption{Effect of noise ($\sigma = 0.5, 1.0, 1.5$) and overlap on metrics as a function of $\varepsilon$}\label{fig:noise}
\end{figure*}

Thus, with this layer-specific metric for each layer, we can use the same value of $\varepsilon_0$ for each layer of interest. Due to activation functions, which are applied after the affine projection, the bound obtained by using weighted distance is still going to be conservative in several directions. Moreover, computation of eigenvalue decomposition for large weight matrices is also intractable. Thus, we provide an alternate mechanism to compute the values of $\bar{\varepsilon}$ parameters. BO is proposed for selecting the value of $\bar{\varepsilon} = \{\varepsilon_i: i \in I \}$ for each layer. BO is computationally tractable for most situations as parameter tuning for Epistemic classifiers does not require base model retraining and repetition of ball-tree construction. Moreover, layers of interest are chosen to be very few ($1-3$), so the number of parameters to jointly tune is very limited. The choice of layers where support is constructed is not considered as a hyperparameter for optimization. If we consider only output layer, then classifier will lack robustness against adversarial perturbations as the neighborhood might already be consistent with targeted class. If we consider only input layer, then it might be computationally expensive due to data dimensionality and in some cases we might lose some robustness as semantic similarity in input space might not be well-represented by~$\ell_p$-norms. Moreover, if neighborhood is pure in or close to input layer and is also pure in or close to final layer, and these neighbors are consistent, then the benign/malign perturbation was not successful in leading to misclassification or incorrect belief. Thus, a combination of layer closer to input, where search is tractable, and layer closer to output are considered. We will further justify this choice with experimental results. 

During inference stage, a testing sample $x$ is used with the Epistemic classifier according to Algorithm~\ref{alg:EClassifier_Inference}. The activations due to the sample $x$ from layers of interest $I$ are extracted. Epistemic classifier then finds the neighborhood and support for the testing input $x$ across the selected layers based on the extracted activation values. This operation to find support in layer $i$ is denoted by operator $\Psi$, which uses activation from test point~$\Gamma_i^x$ in layer $i$ and neighborhood operator $\mathcal{N}_i$. Justification is then created using~\eqref{eq:justification} with support from layers of interest. This justification and belief is then used to obtain justified belief.

\begin{table*}[!h]
	\centering
	\begin{tabular}{|ll|rrr|rrr|rrr|rrr|}
		\toprule
		&           & \multicolumn{3}{c|}{Nominal} & \multicolumn{3}{c|}{Gaussian} & \multicolumn{3}{c|}{Uniform} & \multicolumn{3}{c|}{Large-Noise} \\
		&           &     $F_{IK}$  &     $A_{IK}$ &     $A_{\lnot IK}$ &       $F_{IK}$  &     $A_{IK}$ &     $A_{\lnot IK}$ &      $F_{IK}$  &     $A_{IK}$ &     $A_{\lnot IK}$ &     $F_{IK}$  &     $A_{IK}$ &     $A_{\lnot IK}$ \\ \midrule
		Grid& Base &  0.661 &  0.995 &  0.826 &    0.720 &  0.863 &  0.573 &   0.770 &  0.721 &  0.557 &         0.933 &  0.433 &  0.494 \\
		model:& $k$-NN &  0.692 &  0.990 &  0.818 &    0.720 &  0.869 &  0.560 &   0.754 &  0.748 &  0.486 &         0.776 &  0.457 &  0.368 \\
		FC  & $\varepsilon$-NN &  0.615 &  0.992 &  0.850 &    0.532 &  0.858 &  0.696 &   0.331 &  0.738 &  0.656 &         \textbf{0.001} &  0.750 &  0.437 \\
		(93.74)& H-2&  0.581 &  0.995 &  0.858 &    0.494 &  0.878 &  0.688 &   0.301 &  0.757 &  0.651 &         \textbf{0.001} &  0.750 &  0.437 \\ \midrule
		
		Iris& Base &  0.889 &  1.000 &  1.000 &    0.867 &  1.000 &  1.000 &   0.822 &  0.973 &  1.000 &         0.800 &  0.528 &  0.222 \\
		model:& $k$-NN &  0.911 &  1.000 &  1.000 &    0.889 &  1.000 &  1.000 &   0.889 &  0.975 &  1.000 &         0.822 &  0.459 &  0.500 \\
		FC   & $\varepsilon$-NN &  0.889 &  1.000 &  1.000 &    0.800 &  1.000 &  1.000 &   0.800 &  1.000 &  0.889 &         \textbf{0.178} &  0.500 &  0.459 \\
		(100)& H-2 &  0.889 &  1.000 &  1.000 &    0.800 &  1.000 &  1.000 &   0.800 &  1.000 &  0.889 &         \textbf{0.178} &  0.500 &  0.459 \\ \midrule
		
		Italy & Base &  0.897 &  0.985 &  0.689 &    0.862 &  0.983 &  0.662 &   0.814 &  0.956 &  0.665 &         0.796 &  0.712 &  0.548 \\
		model:& $k$-NN &  0.909 &  0.981 &  0.691 &    0.864 &  0.983 &  0.657 &   0.826 &  0.954 &  0.654 &         0.718 &  0.717 &  0.579 \\
		CNN & $\varepsilon$-NN &  0.894 &  0.977 &  0.761 &    0.821 &  0.970 &  0.793 &   0.588 &  0.942 &  0.844 &        \textbf{ 0.001 }&  1.000 &  0.678 \\
		(95.40)& H-2 &  0.856 &  0.982 &  0.791 &    0.765 &  0.986 &  0.785 &   0.543 &  0.957 &  0.836 &         \textbf{0.001} & 1.000 &  0.678 \\
		\midrule
		&           & \multicolumn{3}{c|}{Nominal} & \multicolumn{3}{c|}{Gaussian} & \multicolumn{3}{c|}{Uniform} & \multicolumn{3}{c|}{Adv-Noise} \\
		&           &     $F_{IK}$  &     $A_{IK}$ &     $A_{\lnot IK}$ &       $F_{IK}$  &     $A_{IK}$ &     $A_{\lnot IK}$ &      $F_{IK}$  &     $A_{IK}$ &     $A_{\lnot IK}$ &     $F_{IK}$  &     $A_{IK}$ &     $A_{\lnot IK}$ \\
		\midrule
		SynCon& Base &  0.907 &  0.996 &  0.929 &    0.730 &  0.995 &  0.728 &   0.727 &  0.977 &  0.756 &  0.243 &  0.562 &  0.577 \\
		model:& $k$-NN &  0.907 &  1.000 &  0.893 &    0.733 &  0.991 &  0.738 &   0.730 &  0.977 &  0.753 &  \textbf{0.440} &  0.955 &  0.274 \\
		CNN & $\varepsilon$-NN &  0.897 &  1.000 &  0.903 &    0.493 &  0.980 &  0.868 &   0.433 &  0.985 &  0.865 &  0.343 &  0.942 &  0.381 \\
		(99.00)& H-2 &  0.863 &  1.000 &  0.927 &    0.437 &  0.992 &  0.870 &   0.393 &  0.983 &  0.874 &  0.307 &  0.957 &  0.404 \\ \midrule
		
		MNIST & Base &  0.622 &  0.998 &  0.899 &    0.424 &  0.988 &  0.738 &   0.306 &  0.888 &  0.491 &  0.851 &  0.026 &  0.011 \\
		model:& $k$-NN &  0.671 &  0.999 &  0.881 &    0.449 &  0.996 &  0.720 &   0.234 &  0.976 &  0.501 &  0.222 &  0.028 &  0.023 \\
		FC   & $\varepsilon$-NN &  0.590 &  0.997 &  0.907 &    0.371 &  0.982 &  0.763 &   0.108 &  0.901 &  0.577 &  \textbf{0.000} &  0.000 &  0.024 \\
		(96.03)& H-2 &  0.495 &  0.999 &  0.922 &    0.279 &  0.997 &  0.785 &   0.071 &  0.982 &  0.584 &  \textbf{0.000} &  0.000 &  0.024 \\ \midrule	
		
		GTSRB & Base &  0.466 &  1.000 &  0.952 &    0.475 &  1.000 &  0.913 &   0.485 &  1.000 &  0.833 &  0.631 &  0.064 &  0.112 \\
		model:& $k$-NN &  0.595 &  1.000 &  0.936 &    0.574 &  1.000 &  0.892 &   0.535 &  0.999 &  0.816 &  0.414 &  0.099 &  0.069 \\
		CNN  & $\varepsilon$-NN &  0.488 &  0.999 &  0.950 &    0.488 &  0.998 &  0.912 &   0.485 &  0.996 &  0.837 &  \textbf{0.140} &  0.015 &  0.092 \\
		(97.41)& H-2  &  0.377 &  1.000 &  0.959 &    0.373 &  1.000 &  0.927 &   0.359 &  1.000 &  0.865 &  \textbf{0.102} &  0.018 &  0.089 \\

		\bottomrule
	\end{tabular}
	\caption{Results on Grid Stability, Iris, and Italy Power dataset in top part and results on Synthetic Control (SynCon) time-series dataset as well as MNIST and GTSRB image datasets in bottom part. Base model test accuracy on nominal data is provided in first column.}
	\label{table:result_noise}
\end{table*}

\section{Experiments and Discussion}

In this section, we will first identify metrics for comparison and then describe experiments with our inferences and insights from the results. We introduce the concept of augmented confusion matrix (ACM) that consists of three sub-matrices, where each sub-matrix is a confusion matrix for predicted label versus true label under assertion of IK (top), IMK (middle), and IDK (bottom). See Fig.~\ref{fig:acm} for examples of ACM. Several robust generalization metrics can be devised using this matrix. In this work, we will use coverage or fraction of IK ($F_{IK}$), accuracy over IK samples ($A_{IK}$), and accuracy over non-IK samples ($A_{\lnot IK}$).

To generate Fig.~\ref{fig:acm}, the Epistemic classifier with a fully-connected neural network is trained with data generated using \texttt{make\_blobs} (bivariate Gaussian) from scikit-learn. We use $\varepsilon$-ball neighborhood with $\ell_2$ distance metric and we use ball-tree to conduct search. Given 4 distinct values of $\varepsilon$ on log-scale, the results on test data is shown with ACM in Fig.~\ref{fig:acm}. It is noted that as value of $\varepsilon$ is small (resp. large), we have almost all assertions as IDK (resp. IMK). Moreover, in general, the confusion matrix under small $\varepsilon$ for IDK and that under large $\varepsilon$ for IMK is also similar. As $\varepsilon$ increases the points in IDK region get assigned to IK and some to IMK region. At higher values, points in IK region move to IMK and finally at large $\varepsilon$ all points are assigned to IMK region.

In Fig.~\ref{fig:noise}, we conduct experiment with similar setup but now study variation due to extent of overlap. Irrespective of choice of parameters, we should not be able to provide reliability for inputs coming from region that has overlap, but we should be able to characterize that region of confusion. Here we show the scatter plot of complete dataset and variation in accuracy of IK ($A_{IK}$) and fraction of IK ($F_{IK}$) metrics with choice of $\varepsilon$. Note that maximum value of $F_{IK}$ reduces as overlap increases implying that region where reliable predictions can be made has reduced, e.g. $F_{IK} \approx 1.00$, when noise std. $\sigma = 0.5$ and $F_{IK} \approx 0.44$, when noise std. is $1.5$. 

We then conducted experiments using Grid Stability~\cite{arzamasov2018towards} and Iris dataset from UCI Repository~\cite{Dua:2019}, Italy Power Demand classification~\cite{keogh2006intelligent} and Synthetic Control (SynCon) dataset~\cite{alcock1999time} from UCR Time-series Repository~\cite{UCRArchive}, MNIST image dataset~\cite{lecun1998mnist}, and German Traffic Sign Recognition Benchmark (GTSRB) dataset~\cite{stallkamp2012man} to get prediction reliability under various perturbations for different baseline models. We have used convolutional neural networks (CNN) for Italy Power, SynCon, and GTSRB datasets and for other datasets we have used fully-connected (FC) neural networks as base classifiers. After generating these base classifers, we build Epistemic classifiers using Algorithm~\ref{alg:EClassifier}. We show performance of the Epistemic classifier on normal holdout testing data as well as on test datasets with perturbations to show context change or drift. In top part of Table~\ref{table:result_noise}, we consider Gaussian noise, uniform noise, and large uniform noise to create rubbish and highly overlapping examples. In bottom part of Table~\ref{table:result_noise}, we consider Gaussian noise, uniform noise, and adversarial perturbation using Basic Iterative Method~\cite{kurakin2016adversarial} from Adversarial Robustness Toolbox~\cite{nicolae2018adversarial}. Details of individual datasets, base model architecture, perturbation magnitudes, and base model accuracy are reported in Supplementary material (S2). The baseline approach for comparison uses a calibrated threshold over softmax layer output that allows to abstain when max over softmax layer values is below a certain threshold. To make the comparison fair, we first determine the optimal $\varepsilon$ for $\varepsilon$-NN neighborhood operator, where $F_{IK}$ is maximized. We tune the parameters for our baseline and $k$-NN support such that they have the similar coverage as $\varepsilon$-ball support, then we make comparison in the table to compare accuracy and performance under nominal data and various perturbations. 

Few insights from Table~\ref{table:result_noise} are discussed next. Softmax thresholding baseline performs well on zero-mean small perturbations and is comparable to performance with $k$-NN neighborhood operator. However, specifically consider Grid, MNIST, and GTSRB dataset, under large uniform noise and adversarial perturbation cases, where distance to hyperplane and location of other data points would start to differ, we see that baseline has high fraction of IK but very poor accuracy on IK. Under these cases, ideal solution is to have high $F_{IK}$ and high $A_{IK}$ for adversarial robustness or at least be able to assert IDK and get very low $F_{IK}$ for detection of malicious perturbations. This is true for $\varepsilon$-NN and H-2 neighborhood operator as highlighted in bold. Note that accuracy metric for large perturbation does not matter as the test samples are more likely to be rubbish examples or indeed in vicinity of wrong class. The results also highlight that $\varepsilon$-NN and H-2 cases have lower coverage than $k$-NN in all cases, however it has superior reliability under large perturbations. We have omitted the results for H-1 support as it is similar to $k$-NN operator, but those results are included in supplementary material. In supplementary material (S3), we have provided additional results where we consider neighborhood only in logit layer and in logit and intermediate layers together. Like baseline, support in only logit layer has good coverage, but lacks robustness under perturbations, thus gathering support from at least two layers is preferable. Although, the metrics in Table~\ref{table:result_noise} denote ability of developed approach to provide individual reliability in scenarios of extrapolation (i.e. anomaly detection) and overlap, additional outcomes showing ability to characterize overlap between subset of classes are omitted due to space restrictions.    

This paper deals with the reliability due to observability and noise. While BIM-based adversarial attacks have been considered, we have explored other inference-time and backdoor attacks as well. A detailed treatment of adversarial robustness is being planned for a follow-on paper. 

\subsection{Challenges and Limitations}
In this approach of identifying support for justification, there are five main challenges, which have not been completely addressed in this research yet. Firstly, large dimensionality of input and embedded space in early layers leads the k-d tree and ball-tree search algorithm to be computationally expensive and is impractical for large convolutional layers of state-of-the-art models. Locality-sensitive hashing~\cite{slaney2008locality} and other approximate neighbor search approaches~\cite{andoni2006near} can be used to alleviate these issues. Secondly, number of training data points, i.e., $|X|$ leads to $O(\log|X|)$ average case complexity for search and $O(|X|)$ for space, which becomes of concern for large datasets. Thirdly, the most crucial challenge that we have come across is the closeness of semantic distance between points and the mathematically-convenient $\ell_2$-norm distance metric in input and early layers. The closeness of these metrics increases as we go deeper in the network, since neural network training is expected to bring data points from same classes closer to enable linear separation, however in input and initial layers this closeness is usually not available. Fourthly, sparsity of data in neighborhood either due to high dimensionality or low sample size will also adversely affect the performance of this classifier. Finally, the current approach is suited only for neural networks, a model-agnostic approach using suitable input representation and output with Platt scaling~\cite{platt1999probabilistic} will be explored for other ML models. These challenges will be further explored in near future.    

\section{Conclusion}
In this work, we have introduced a novel approach to provide individual reliability of predictions from ML models. We leveraged the notion of Knowledge as Justified Belief from field of epistemology to create Epistemic classifiers in ML. Epistemic classifier adds more contextual information based on location of training data points in input and hidden layers to add reliability on individual predictions, wherever plausible. We also introduced various domain-agnostic neighborhood operators which can be used to gather evidence to construct justification. We analyzed the effect of parameter choices, noise in training dataset, and benign/malign perturbations on performance, where performance was reported using new metrics relevant to the epistemic analysis. We contrasted the performance of Epistemic classifiers with a baseline of thresholding softmax layer values and demonstrated superiority in robustness. The results were demonstrated on simulated and real datasets with few features, time-series data, and image data to show flexibility of the solution. Finally, key limitations were highlighted which will pave the way for future research. 

\section{Acknowledgments}
This work is a part of the GE Humble AI initiative. Authors would like to thank the reviewers and senior program committee members for helpful reviews.

\bibliographystyle{aaai}
\bibliography{confs}

\section{Supplementary Material}

\section{(S1) Analysis of $\varepsilon$ and metric}

\begin{rem}[$\varepsilon$ bound]
If~$W_i \in \mathbb{R}^{m \times n}$ is weight matrix of the layer $i$ with input dimension $m$ and output dimension $n$, $\lambda_i^*$ is the largest eigenvalue of the matrix $W_{i}W_{i}^T$, $L_i$ is the Lipschitz constant of the activation function of layer $i$, then given $\varepsilon_i$ for layer $i$, the value of $\varepsilon_{i+1}$ is bounded as follows:
$$\varepsilon_{i+1} \leq  L_i \sqrt{\lambda_i^*} \varepsilon_i.$$ 
\end{rem}
\begin{proof}
Defined an input to the layer $i$ as $x \in \mathbb{R}^{m \times 1}$, output of the layer $i$ given the input $x$ is defined as $h_i(x) = \phi_i(W_i^Tx + b_i)$, where $b_i \in \mathbb{R}^{n \times 1}$ and $\phi_i : \mathbb{R}^{n \times 1} \rightarrow \mathbb{R}^{n \times 1}$ is the activation function for the layer $i$. Given another input $z \in \mathbb{R}^{m \times 1}$ where $(x - z)^T(x-z) \leq \varepsilon_i^2$, and assuming Lipschitz constant for layer $i$ activation function $\phi_i(.)$ is $L_i$, then we have
\begin{equation*}
\begin{aligned}
&\left\Vert\ h_i(z) - h_i(x) \right\Vert_2
\\
&=\left\Vert\ \phi_i(W_i^Tz+b_i) - \phi(W_i^Tx+b_i) \right\Vert_2
\\
&\leq \max_{z} \left\Vert\ L_i(W_i^Tz+b_i) - L_i(W_i^Tx+b_i) \right\Vert_2
\\
& = \max_{z} \left\Vert\ L_iW_i^T(x - z) \right\Vert_2
\\
& = \max_{z}  \sqrt{L_i^2(x - z)^TW_{i}W_{i}^T(x-z)} 
\\
\end{aligned}
\end{equation*}
To solve the upper bound, let's define $\delta = x - z$ and ignore the Lipschitz constant for now, then we can reformulate the problem as
\begin{equation*}
\label{eq:t}
\begin{aligned}
\max_{\delta}  \delta^TW_{i}W_{i}^T\delta 
\\
\text{st } \delta^T\delta \leq \varepsilon_i^2
\end{aligned}
\end{equation*}
Lagrangian function of the formulation is
\begin{equation*}
\begin{aligned}
\mathcal{L}(\delta,\lambda) = \delta^TW_{i}W_{i}^T\delta  - \lambda(\delta^T\delta-\varepsilon_{i}^2) 
\end{aligned}
\end{equation*}
Solution for the Lagrangian function is
\begin{equation*}
\label{eq:t}
\begin{aligned}
\nabla_{\delta} \mathcal{L} &= 2 W_{i}W_{i}^T\delta -2\lambda\delta = 0
\\
\nabla_{\lambda} \mathcal{L} &= \delta^{T}\delta - \varepsilon_i^2 = 0
\end{aligned}
\end{equation*}
\\
Thus, we have 
\begin{equation*}
\label{eq:t}
\begin{aligned}
&\delta^{*T} W_{i}W_{i}^T\delta^* = \lambda \delta^{*T}\delta^* = \lambda \varepsilon_{i}^2 
\\
&\left\Vert\ h_{i}(z) - h_{i}(x) \right\Vert_2 \leq {L_i}\sqrt{\lambda^*} \varepsilon_{i}
\end{aligned}
\end{equation*}
where $\lambda^*$ is the largest eigenvalue of the matrix $W_iW_i^T$.
Thus, relation of $\varepsilon_{i}$ across consecutive layers is given by: $$\varepsilon_{i+1} \leq L_{i}\sqrt{\lambda^*} \varepsilon_i.$$
Considering the output of $i_{th}$ layer given input from the first layer as
\begin{equation*}
\label{eq:t}
\begin{aligned}
\tilde{h}_i(x) = h_{i-1}\left(h_{i-2}\left(... h_0\left(x\right)\right)\right).
\end{aligned}
\end{equation*}
Similarly, we know that for $(x - z)^T(x-z) \leq \varepsilon^2$, we have
\begin{equation}
\label{eq:t}
\begin{aligned}
&\left\Vert\ \tilde{h}_i(z) - \tilde{h}_i(x) \right\Vert_2 \notag
\\
&\leq \max_{z} \left\Vert\  \prod_{j=0}^{i-1} L_j W_{i-1}^TW_{i-2}^T...W_{0}^T (x - z) \right\Vert_2
\\
&\leq L^{i} \sqrt{\tilde{\lambda}^*} \varepsilon,
\end{aligned}
\end{equation}
where $\tilde{\lambda}^*$ is the largest eigenvalue of the matrix $W_0W_1...W_{i-1}W_{i-1}^T...W{_1}^TW{_0}^T$.
\end{proof}

However, this bound might be loose in terms of direction of the vector $\delta = x-z$. Therefore, alternate approach is to derive a distance metric for each layer. 

\begin{rem}[Bound with weighted distance - new version]\label{rem:distance}
Let $\widetilde{W_i} = W_1W_2...W_{i}$, $\Lambda = \text{diag}([\lambda_1, \lambda_2,...\lambda_n])$ and $V = [v_1, v_2,...v_n]$ as the eigenvalues and corresponding eigenvectors of matrix $\widetilde{W_i}^T\widetilde{W_{i}}$, and $d_0$ is $\ell_2$ distance metric in input space. If $d_0(x, z) \leq \varepsilon_0 $ then the distance function matrix $D_i$ for neighbor search in any layer $i$:
\begin{equation}
\label{eq:t}
\begin{aligned}
D_i = V\Lambda^{-1}V^T
\end{aligned}
\end{equation}
satisfies the following relation:	$$\sqrt{(h_i(x)- h_i(z))^T D_i (h_i(x)- h_i(z))} \leq \varepsilon_0$$
\end{rem}
\begin{proof}
Assume activation functions are ReLu and Lipschitz constant is 1. Defined the following metric $d:X\times X \rightarrow \mathbb{R}$, where, for each layer i, we have $d_i(x,z) = \sqrt{(x-z)^TD_i(x-z)}$ and $D_i$ is a symmetric PSD matrix. For simplification, we start with $h(x) = W^Tx + b$ where $x \in \mathbb{R}^{m \times 1}$, $W \in \mathbb{R}^{m \times n}$ and $b \in \mathbb{R}^{n \times 1}$. We make an assumption that $n \leq m$, which is a common practice since most of neural networks have less neurons in deeper layers. Using SVD, we can write $W=U\Sigma V^T$. Define $D = V\Sigma^+{\Sigma^T}^+V^T = V\Lambda^{-1}V^T$, where $\Sigma^+$ is pseudoinverse of $\Sigma$, we have

\begin{equation*}
\begin{aligned}
&(h(x)-h(z))^TD(h(x)-h(z))
\\
&=(x-z)^TWDW^T(x-z)
\\
&=(x-z)^T U\Sigma V^T D V\Sigma^T U^T(x-z)
\\
&=(x-z)^T U\Sigma V^T V\Sigma^+{\Sigma^T}^+V^T V\Sigma^T U^T(x-z)
\\
&\leq (x-z)^TUU^T(x-z)
\\
&= (x-z)^T(x-z)
\end{aligned}
\end{equation*}
Additionally, if $(x - z)^T(x-z) \leq {\varepsilon_0}^2$, then we have 
\begin{equation*}
\begin{aligned}
\sqrt{(h(x)- h(z))^T D (h(x)- h(z))} \leq \varepsilon_0.
\end{aligned}
\end{equation*}
\end{proof}

\begin{rem}[Bound with weighted distance - previous version]\label{rem:distance}
Let $\widetilde{W_i} = W_1W_2...W_{i}$, $\Lambda = \text{diag}([\lambda_1, \lambda_2,...\lambda_n])$ and $U = [v_1, v_2,...v_n]$ as the first $n$ non-zero eigenvalues and corresponding eigenvectors of matrix $\widetilde{W_i}\widetilde{W_{i}}^T$, and $d_0$ is $\ell_2$ distance metric in input space. If $d_0(x, z) \leq \varepsilon_0 $ then the distance function matrix $D_i$ for neighbor search in any layer $i$:
\begin{equation}
\label{eq:t}
\begin{aligned}
D_i = \widetilde{W_i}^TU\Lambda^{-1}U^T\widetilde{W_{i}}
\end{aligned}
\end{equation}
satisfies the following relation:	$$\sqrt{(h_i(x)- h_i(z))^T D_i (h_i(x)- h_i(z))} \leq \varepsilon_0$$
\end{rem}
\begin{proof}
Assume activation functions are ReLu and Lipschitz constant is 1. Defined the following metric $d:X\times X \rightarrow \mathbb{R}$, where, for each layer i, we have $d_i(x,z) = \sqrt{(x-z)^TD_i(x-z)}$ and $D_i$ is a symmetric PSD matrix. For simplification, we start with $h(x) = W^Tx + b$ where $x \in \mathbb{R}^{m \times 1}$, $W \in \mathbb{R}^{m \times n}$ and $b \in \mathbb{R}^{n \times 1}$. We make an assumption that $n \leq m$, which is a common practice since most of neural networks have less neurons in deeper layers. Based on previous derivation, we know that distance between x and z will be local maximum after projection if direction of the vector $\delta = x - z$ in the input space is aligned with one of the eigenvectors of the matrix $WW^T$. Assuming there are n non-zero eigenvalues for the matrix $WW^T$ since we know $\text{rank}(WW^T) \leq n$ and Lipschitz constant of activation~$L_i$ is~$1$. Define $\Lambda = \text{diag}([\lambda_1, \lambda_2,...\lambda_n]) \in \mathbb{R}^{n \times n}$ and $U = [v_1, v_2,...v_n] \in \mathbb{R}^{m \times n}$ as first n non-zero eigenvalues and corresponding eigenvectors, we could obtain the following metric for neighbor search:
\begin{equation*}
\begin{aligned}
D = W^TU\Lambda^{-1}U^TW.
\end{aligned}
\end{equation*}
Additionally, if $(x - z)^T(x-z) \leq {\varepsilon_0}^2$, then we have 
\begin{equation*}
\begin{aligned}
\sqrt{(h(x)- h(z))^T D (h(x)- h(z))} \leq \varepsilon_0.
\end{aligned}
\end{equation*}

Similarly, for $i^{th}$ layer of a fully-connected deep neural network, we have
\begin{equation*}
\begin{aligned}
D_i = W_{i-1}^T...W_{1}^TW_{0}^TU\Lambda^{-1}U^TW_0W_1...W_{i-1}.
\end{aligned}
\end{equation*}
\end{proof}

\section{(S2) Dataset and Base Model Description}
Performance of the base classifiers on the original test data and their architectures are listed as follow:

\textbf{Fully connected neural network:}

UCI-Grid stability ~\cite{Dua:2019}~\cite{arzamasov2018towards} (FNN): 

Base classifier accuracy: 93.74\% 

Input dimension: 13 

Network Architecture: 

Dense layer: 32 + 32, Output: 2

Activation: ReLu

Perturbation: 

Gaussian: 0.03 standard deviation of training data

Uniform noise range: +/- 0.09 standard deviation of training data

Large perturbation: Uniform +/- 0.5*(max-min)

Selected layer: 2nd Dense layer + logit\\

UCI-IRIS ~\cite{Dua:2019} (FNN): 

Base classifier accuracy: 100\% 

Input dimension: 4

Network Architecture: 

Dense layer: 8 + 5, Output: 3

Activation: ReLu

Perturbation: 

Gaussian: 0.03 standard deviation of training data

Uniform: +/- 0.09 standard deviation of training data

Large perturbation: Uniform +/-0.5*(max-min)

Selected layer: 2nd Dense layer + logit\\

MNIST ~\cite{lecun1998mnist} (FNN), Data scale to [0,1]:  

Base classifier accuracy: 96.03\% 

Input dimension: 784

Network Architecture: 

Dense layer: 32 + 32, Output: 10

Activation: ReLu

Perturbation: 

Gaussian: 0.1 standard deviation of training data

Uniform: +/- 0.3 standard deviation of training data

Adversarial perturbation: +/- 0.2

Selected layer: 2nd Dense layer + logit\\

\textbf{Convolutional neural network:}

UCR-Italy Power Demand ~\cite{keogh2006intelligent} (CNN): 

Base classifier accuracy: 95.4\% 

Input dimension: 1 * 24

Network Architecture: 

Convolutional layer (num, kernel size 1, kernel size 2): (6,1,4) + (16,1,12) + (8, 1, 6) + Global average pooling (GAP), Output: 2

Activation: ReLu

Perturbation:

Gaussian: 0.2 standard deviation of training data

Uniform: +/- 0.6 standard deviation of training data

Large perturbation: Uniform +/-0.5*(max-min)

Selected layer: 3rd Convolutional layer + GAP\\

UCR-SynCon ~\cite{alcock1999time} (CNN): 

Base classifier accuracy: 99\% 

Input dimension: 1 * 60

Network Architecture: 

Convolutional layer (num, kernel size 1, kernel size 2): (9,1,12) + (12,1,16) + (9,1,12) + Global average pooling (GAP), Output: 6

Activation: ReLu

Perturbation:

Gaussian: 0.5 standard deviation of training data

Uniform: +/- 1 standard deviation of training data

Adversarial perturbation: +/- 0.2

Selected layer: 3rd Convolution layer + GAP layer\\

GTSRB \cite{stallkamp2012man} (CNN), Data scale to [0,1]: 

Base classifier accuracy: 97.41\%  

Input dimension: 32*32*3

Network Architecture: 

Conv(6,5,5) + maxPool((2,2), (2,2)) + Conv(16,5,5) + maxPool((2,2), (2,2)) + Conv(64,5,5) + Global average pooling (GAP) + Dense layer 32, Output: 8

Activation: ReLu

Perturbation:

Gaussian: 0.033 standard deviation of training data

Uniform: +/- 0.1 standard deviation of training data

Adversarial perturbation: +/- 0.05

Selected layer: GAP layer + logit

\textbf{Preprocessing for GTSRB Data:} Training, validation and testing data for the traffic sign classification are generated from GTSRB traffic sign dataset. We group traffic sign types into eight classes which are: speed limit, no passing, end of restriction, warning, priority, yield, stop and direction. In the end, we have 34799 images for training, 4410 for validation and 12630 images for testing. 

\begin{figure}[!h]
	\centering
	\begin{minipage}[b]{0.5\textwidth}
		\includegraphics[width=\textwidth]{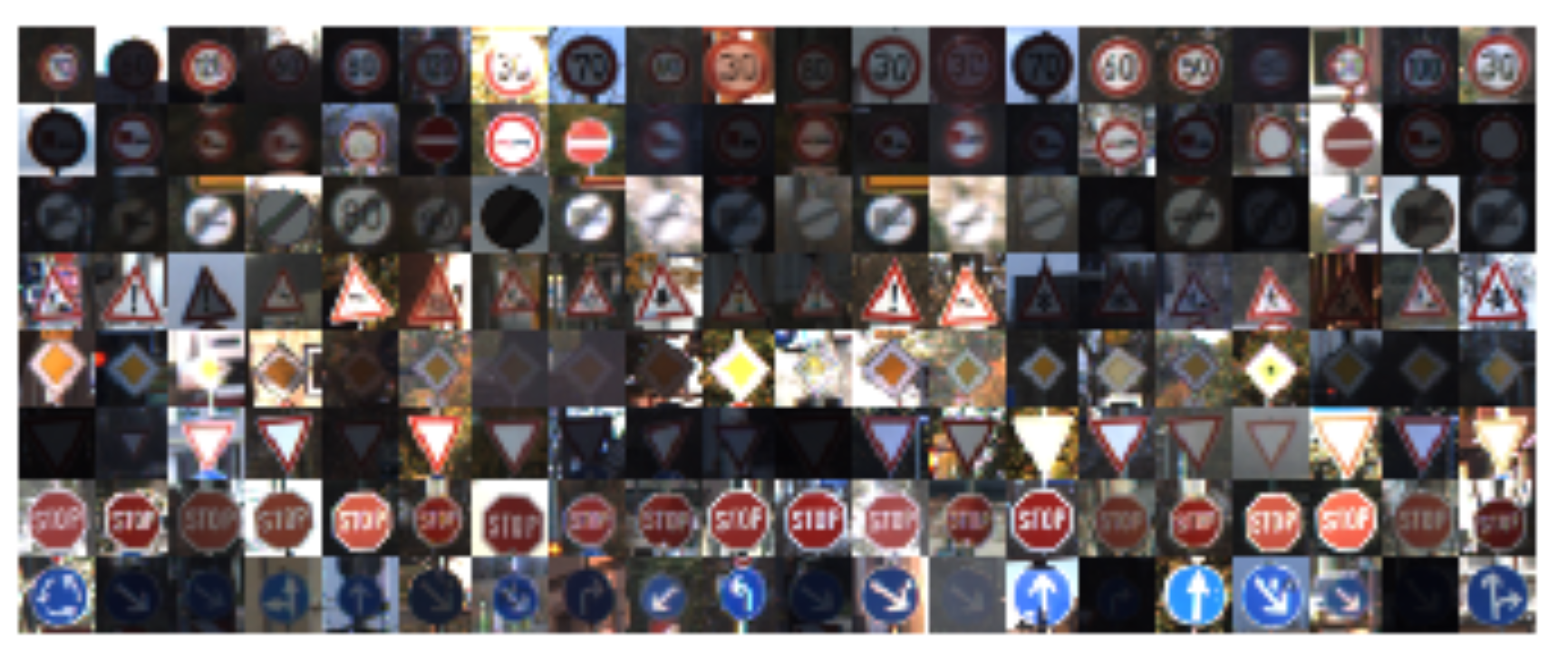}
	\end{minipage}  
	\caption{Traffic sign dataset, from top to bottom are: speed limit, no passing, end of restriction, warning, priority, yield, stop and direction.}
	\label{fig:s_a}
\end{figure}

\section{(S3) Additional Results}
Table~\ref{table:result_attack_complete} show results for MNIST, SynCon, and GTSRB datasets, where support was constructed in logit layer only and in logit layer and another intermediate layer.
\begin{table*}[!tb]
	\centering
	\begin{tabular}{|ll|rrr|rrr|rrr|rrr|}
		\toprule
		&           & \multicolumn{3}{c|}{Nominal} & \multicolumn{3}{c|}{Gaussian} & \multicolumn{3}{c|}{Uniform} & \multicolumn{3}{c|}{Adv-Noise} \\
		&           &     $F_{IK}$  &     $A_{IK}$ &     $A_{\lnot IK}$ &       $F_{IK}$  &     $A_{IK}$ &     $A_{\lnot IK}$ &      $F_{IK}$  &     $A_{IK}$ &     $A_{\lnot IK}$ &     $F_{IK}$  &     $A_{IK}$ &     $A_{\lnot IK}$ \\
		\midrule
		SynCon& Base &  0.907 &  0.996 &  0.929 &    0.730 &  0.995 &  0.728 &   0.727 &  0.977 &  0.756 &  0.243 &  0.562 &  0.577 \\
	   Model: & KNN &  0.907 &  1.000 &  0.893 &    0.733 &  0.991 &  0.738 &   0.730 &  0.977 &  0.753 &  0.440 &  0.955 &  0.274 \\
		CNN & $\varepsilon$-NN &  0.897 &  1.000 &  0.903 &    0.493 &  0.980 &  0.868 &   0.433 &  0.985 &  0.865 &  0.343 &  0.942 &  0.381 \\
		Accuracy: & H-1 &  0.933 &  1.000 &  0.850 &    0.783 &  0.983 &  0.708 &   0.757 &  0.982 &  0.712 &  0.457 &  0.942 &  0.264 \\
		0.9900 & H-2 &  0.863 &  1.000 &  0.927 &    0.437 &  0.992 &  0.870 &   0.393 &  0.983 &  0.874 &  0.307 &  0.957 &  0.404 \\ \cline{1-14}\hline
		
		& Base &  0.940 &  0.996 &  0.889 &    0.827 &  0.980 &  0.654 &   0.810 &  0.975 &  0.667 &  0.403 &  0.579 &  0.570 \\
		& KNN &  0.953 &  1.000 &  0.786 &    0.840 &  0.976 &  0.646 &   0.827 &  0.968 &  0.673 &  0.510 &  0.745 &  0.395 \\
		SynCon  &  $\varepsilon$-NN &  0.920 &  0.996 &  0.917 &    0.740 &  0.955 &  0.833 &   0.613 &  0.951 &  0.862 &  0.480 &  0.667 &  0.487 \\
		(logit) & H-1 &  0.957 &  0.997 &  0.846 &    0.900 &  0.959 &  0.600 &   0.850 &  0.953 &  0.711 &  0.603 &  0.707 &  0.370 \\
		& H-2 &  0.907 &  1.000 &  0.893 &    0.680 &  0.975 &  0.812 &   0.587 &  0.972 &  0.839 &  0.377 &  0.699 &  0.497 \\ \cline{1-14}\hline
		
		MNIST& Base &  0.622 &  0.998 &  0.899 &    0.424 &  0.988 &  0.738 &   0.306 &  0.888 &  0.491 &  0.851 &  0.026 &  0.011 \\
		Model:& KNN &  0.671 &  0.999 &  0.881 &    0.449 &  0.996 &  0.720 &   0.234 &  0.976 &  0.501 &  0.222 &  0.028 &  0.023 \\
		FC & $\varepsilon$-NN &  0.590 &  0.997 &  0.907 &    0.371 &  0.982 &  0.763 &   0.108 &  0.901 &  0.577 &  0.000 &  0.000 &  0.024 \\
		Accuracy: & H-1 &  0.696 &  0.997 &  0.876 &    0.523 &  0.986 &  0.689 &   0.270 &  0.944 &  0.489 &  0.222 &  0.028 &  0.023 \\
		0.9603 & H-2 &  0.495 &  0.999 &  0.922 &    0.279 &  0.997 &  0.785 &   0.071 &  0.982 &  0.584 &  0.000 &  0.000 &  0.024 \\ \cline{1-14}\hline
		
		& Base &  0.814 &  0.995 &  0.809 &    0.598 &  0.973 &  0.653 &   0.466 &  0.814 &  0.436 &  0.909 &  0.025 &  0.010 \\
		& KNN &  0.815 &  0.995 &  0.808 &    0.589 &  0.980 &  0.649 &   0.382 &  0.871 &  0.452 &  0.325 &  0.026 &  0.023 \\
		MNIST & $\varepsilon$-NN &  0.792 &  0.986 &  0.862 &    0.610 &  0.934 &  0.703 &   0.358 &  0.736 &  0.543 &  0.000 &  0.000 &  0.024 \\
		(logit) & H-1 &  0.859 &  0.986 &  0.803 &    0.708 &  0.940 &  0.612 &   0.515 &  0.777 &  0.437 &  0.325 &  0.026 &  0.023 \\
		& H-2 &  0.724 &  0.995 &  0.868 &    0.481 &  0.982 &  0.716 &   0.225 &  0.870 &  0.537 &  0.000 &  0.000 &  0.024 \\ \cline{1-14}\hline
		
	   GTSRB& Base &  0.466 &  1.000 &  0.952 &    0.475 &  1.000 &  0.913 &   0.485 &  1.000 &  0.833 &  0.631 &  0.064 &  0.112 \\
     Model:& KNN &  0.595 &  1.000 &  0.936 &    0.574 &  1.000 &  0.892 &   0.535 &  0.999 &  0.816 &  0.414 &  0.099 &  0.069 \\
     CNN & $\varepsilon$-NN &  0.488 &  0.999 &  0.950 &    0.488 &  0.998 &  0.912 &   0.485 &  0.996 &  0.837 &  0.140 &  0.015 &  0.092 \\
     Accuracy: & H-1 &  0.584 &  0.999 &  0.939 &    0.586 &  0.998 &  0.892 &   0.592 &  0.996 &  0.794 &  0.452 &  0.091 &  0.073 \\
     0.9741 & H-2 &  0.377 &  1.000 &  0.959 &    0.373 &  1.000 &  0.927 &   0.359 &  1.000 &  0.865 &  0.102 &  0.018 &  0.089 \\ \cline{1-14}\hline
		
		& Base &  0.740 &  0.999 &  0.903 &    0.741 &  0.999 &  0.826 &   0.730 &  0.997 &  0.687 &  0.888 &  0.069 &  0.178 \\
     & KNN &  0.737 &  1.000 &  0.903 &    0.716 &  0.999 &  0.842 &   0.679 &  0.991 &  0.750 &  0.555 &  0.085 &  0.077 \\
     GTSRB & $\varepsilon$-NN  &  0.737 &  0.995 &  0.916 &    0.740 &  0.991 &  0.849 &   0.729 &  0.974 &  0.752 &  0.186 &  0.025 &  0.095 \\
     (logit) & H-1 &  0.819 &  0.996 &  0.877 &    0.827 &  0.992 &  0.773 &   0.824 &  0.977 &  0.620 &  0.589 &  0.081 &  0.082 \\
     & H-2 &  0.625 &  1.000 &  0.932 &    0.618 &  0.999 &  0.882 &   0.581 &  0.992 &  0.805 &  0.153 &  0.028 &  0.091 \\
		
		\bottomrule
	\end{tabular}
	\caption{Results on Synthetic Control (SynCon) time-series dataset as well as MNIST and Traffic Sign image datasets}
	\label{table:result_attack_complete}
\end{table*}

\end{document}